\documentclass[10pt,twocolumn,letterpaper]{article}
\usepackage{amsmath}
\usepackage{amssymb}
\usepackage{times}
\usepackage{bm}
\usepackage{xargs}
\usepackage{multirow}
\usepackage{amsfonts}
\usepackage{color}
\usepackage{booktabs}

\newcommandx\includeImageLineWidth[2][1=1.0]{\includegraphics[width=#1\linewidth]{#2}}

\usepackage{array}
\newcommand{\PreserveBackslash}[1]{\let\temp=\\#1\let\\=\temp}
\newcolumntype{C}[1]{>{\PreserveBackslash\centering}p{#1}}
\newcolumntype{R}[1]{>{\PreserveBackslash\raggedleft}p{#1}}
\newcolumntype{L}[1]{>{\PreserveBackslash\raggedright}p{#1}}

\usepackage{wacv}              
\usepackage[accsupp]{axessibility}
\usepackage[utf8]{inputenc} 
\usepackage[T1]{fontenc}    
\usepackage{url}            
\usepackage{booktabs}       
\usepackage{amsfonts}       
\usepackage{nicefrac}       
\usepackage{microtype}      
\usepackage{xcolor}         
\usepackage{algorithm}
\usepackage{algorithmic}
\usepackage{subfigure}
\usepackage{subfiles}
\usepackage{enumitem}
\usepackage{graphicx}
\usepackage{enumitem}
\usepackage{wrapfig}
\usepackage{capt-of}
\usepackage{pifont}

\definecolor{LightCyan}{rgb}{0.9059,0.9961,1}
\usepackage{multirow}
\usepackage{graphicx}
\usepackage[font=small,aboveskip=0pt,belowskip=0pt]{caption}   
\usepackage{makecell}
\usepackage{tabulary}
\definecolor{demphcolor}{RGB}{144,144,144}

\definecolor{mygray}{gray}{0.4}
\newlength\savewidth

\makeatletter\renewcommand\paragraph{\@startsection{paragraph}{4}{\z@}
  {.5em \@plus1ex \@minus.2ex}{-.5em}{\normalfont\normalsize\bfseries}}\makeatother

\usepackage{xspace}

\usepackage[pagebackref,breaklinks,colorlinks]{hyperref}

\usepackage[capitalize]{cleveref}
\crefname{section}{Sec.}{Secs.}
\Crefname{section}{Section}{Sections}
\Crefname{table}{Table}{Tables}
\crefname{table}{Tab.}{Tabs.}

\def\wacvPaperID{1823} 
\def\confName{WACV}
\def\confYear{2024}

\begin{document}

\title{Inflation with Diffusion: Efficient Temporal Adaptation for Text-to-Video Super-Resolution}

\newcommand\Mark[1]{\textsuperscript#1}

\author{Xin Yuan\Mark{1}\thanks{This work has been done during the first author’s internship at Google.}, Jinoo Baek\Mark{2}, Keyang Xu\Mark{2}, Omer Tov\Mark{2}, Hongliang Fei\Mark{2} \\
\Mark{1}University of Chicago \Mark{2}Google \\
{\tt\small yuanx@uchicago.edu
\{jinoo,keyangxu,omertov,hongliangfei\}@google.com}
}
\maketitle

\begin{abstract}
We propose an efficient diffusion-based text-to-video super-resolution (SR) tuning approach that leverages the readily learned capacity of pixel level image diffusion model to capture spatial information for video generation. To accomplish this goal, we design an efficient architecture by inflating the weightings of the text-to-image SR model into our video generation framework. Additionally, we incorporate a temporal adapter to ensure temporal coherence across video frames.  We investigate different tuning approaches based on our inflated architecture and report trade-offs between computational costs and super-resolution quality. 
Empirical evaluation, both quantitative and qualitative, on the Shutterstock video dataset, demonstrates that our approach is able to perform text-to-video SR generation with good visual quality and temporal consistency. To evaluate temporal coherence, we also present visualizations in video format in \href{https://drive.google.com/drive/folders/1YVc-KMSJqOrEUdQWVaI-Yfu8Vsfu_1aO?usp=sharing}{google drive}.
\end{abstract}

\section{Introduction}
\label{sec:intro}
\begin{figure}[h]

\includegraphics[width=\columnwidth]{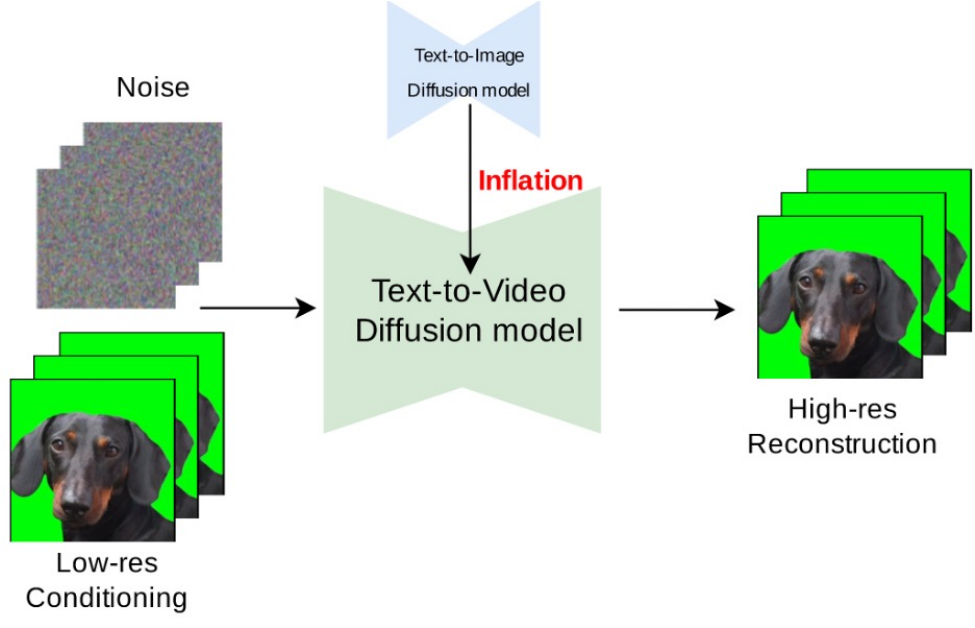}
\centering
\includegraphics[width=0.7\columnwidth]{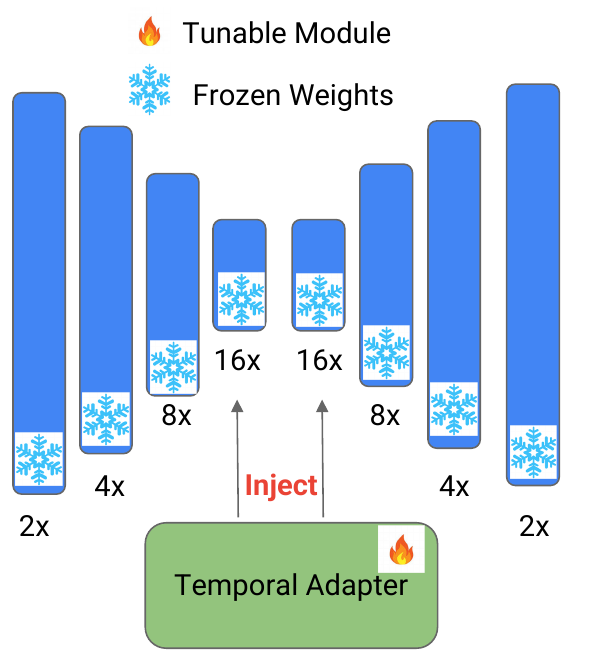}
\caption{Overall architecture of our approach. \textit{Up}: we inflate the UNet weights from a text-to-image model into a text-to-video model to perform a diffusion-based super-resolution task. \textit{Bottom}: we inject and tune a temporal adapter in the inflated architecture while maintaining the UNet weights frozen.
}\label{fig:our_idea}
\vspace{-1.8em}
\end{figure}
Diffusion model~\cite{DBLP:conf/nips/HoJA20,DBLP:conf/iclr/0011SKKEP21}, as a deep generative model, has achieved a new state-of-the-art performance, surpassing GANs~\cite{goodfellow2014generative,
Xu18,
Han17,Karras2019stylegan2} in generative tasks~\cite{DBLP:conf/cvpr/RombachBLEO22,DBLP:conf/nips/SahariaCSLWDGLA22}.
In a diffusion-based text-conditioned generation system,
a base model initially generates a low-resolution image/video, which is subsequently refined by a super-resolution module~\cite{DBLP:conf/nips/SahariaCSLWDGLA22,DBLP:journals/corr/abs-2210-02303, DBLP:conf/iclr/SingerPH00ZHYAG23} to produce high-quality samples. 
Numerous existing diffusion-based text-to-image super-resolution models~\cite{DBLP:conf/nips/SahariaCSLWDGLA22,DBLP:conf/cvpr/RombachBLEO22}, trained on billion-scale text-image dataset, have demonstrated outstanding generation capability.
However, training text-to-video spatial super-resolution is challenging due to the scarcity of high-resolution video data. This scenario motivates the inflation of off-the-shelf image models to video generation tasks~\cite{DBLP:journals/corr/abs-2303-13439,
DBLP:journals/corr/abs-2305-17431, DBLP:journals/corr/abs-2303-12688, DBLP:journals/corr/abs-2212-11565}.
Furthermore, training a video generation model needs exceedingly high computational and memory requirements, which drives techniques that offer cost-effective alternatives to optimize the video models.

Several recently proposed methods~\cite{DBLP:conf/cvpr/BlattmannRLD0FK23} also focus on generating high-quality videos using pretrained latent diffusion models. Temporal attention mechanisms are also commonly used in~\cite{DBLP:journals/corr/abs-2210-02303,DBLP:conf/iclr/SingerPH00ZHYAG23}.
Yet, investigating the trade-offs between video quality and the resource requirements in a fine-tuning stage is not the focus of those works.
~\cite{DBLP:conf/cvpr/BlattmannRLD0FK23} typically requires full tuning of all computational modules to generate high-quality videos, even with pretrained image weights inflated in the video architectures.
In contrast, our approach lies in the applied domain and investigates how tuning efficiency affects the video super-resolution quality. 
More importantly, instead of investigating model inflation in latent space~\cite{DBLP:conf/cvpr/BlattmannRLD0FK23}, our approach is the first to directly work on pixels.
Note that our goal is not to achieve state-of-the-art generation quality. Instead, we aim to establish a practical and efficient tuning system to generate high-resolution videos with reasonable visual quality and temporal consistency.

In this paper, we aim to leverage the readily learned spatial capacity of image weights for efficient and effective text-to-video super-resolution, as shown in the upper of Figure~\ref{fig:our_idea}.
To capture the coherence across video frames, we inject an attention-based temporal adapter into the video architecture. This adapter can be fine-tuned independently while keeping inflated weights frozen, as shown in the bottom of Figure~\ref{fig:our_idea}.
We perform the spatial super-resolution task on the Shutterstock video dataset and validate that our approach is capable of generating videos with good visual quality and temporal consistency. We also demonstrate the trade-off between tuning complexity and generation quality. 

\section{Related Work}
Diffusion-based SR model is conditioned on low-resolution samples, generated by a base generation model, to further produce high-resolution images~\cite{DBLP:conf/nips/SahariaCSLWDGLA22} or videos~\cite{DBLP:journals/corr/abs-2210-02303}. 
With the success of image generation models pre-trained on billion-scale image data, recent research efforts have been made to directly borrow off-the-shelf image models for video tasks. For example, ~\cite{DBLP:journals/corr/abs-2303-13439} load image weights for video generation in a zero-shot manner. ~\cite{DBLP:journals/corr/abs-2305-17431, DBLP:journals/corr/abs-2303-12688, DBLP:journals/corr/abs-2212-11565} adopt model inflation and DDIM~\cite{DBLP:conf/iclr/SongME21} inversion for text-to-video editing. While these studies may not directly apply to video spatial super-resolution task, they provide insightful hints on the feasibility of adopting an image model without necessitating re-training from scratch.

Temporal attention mechanisms that operate on the time axis are commonly adopted in video diffusion approaches~\cite{DBLP:journals/corr/abs-2210-02303,DBLP:conf/iclr/SingerPH00ZHYAG23}. Our method shares the same concept with~\cite{DBLP:conf/cvpr/BlattmannRLD0FK23} in the spirit of borrowing image diffusion models for video generation. However, our approach focus on the applied domain for text-to-video super resolution. More importantly, with facilitating partial tuning of the video architectures, we qualitatively and quantitatively evaluate how different tuning methods affect the generation quality, including visual quality and temporal consistency.

\section{Approach}
\label{sec:method}
Consider a video clip represented as a sequence of $n$ image frames, denoted as $I: [I_1,..., I_n]$, with low spatial resolution $s$, and a text description $t$ for this clip, 
our objective is to generate a new video clip of the same length but with an enhanced resolution $s'$ while preserving the correlation between text and video.  
We aim to exploit the robust spatial understanding of a pre-trained and fixed large-scale image diffusion
model,  repurposing it for video generation that remains temporally consistent. This removes the need for extensive training from scratch on limited high-resolution video data, which is both time- and resource-consuming.
We achieve this goal by inflating the weights of image diffusion model into a video generation architecture (as detailed in Section~\ref{subsec:inflate}) and further tuning an efficient temporal adapter to ensure the continuity and coherence across video frames, discussed in~Section~\ref{subsec:adapter}.
\begin{figure*}[tbh]
\centering
\begin{minipage}[t]{\linewidth}
   \centering
  \includegraphics[width=\columnwidth]{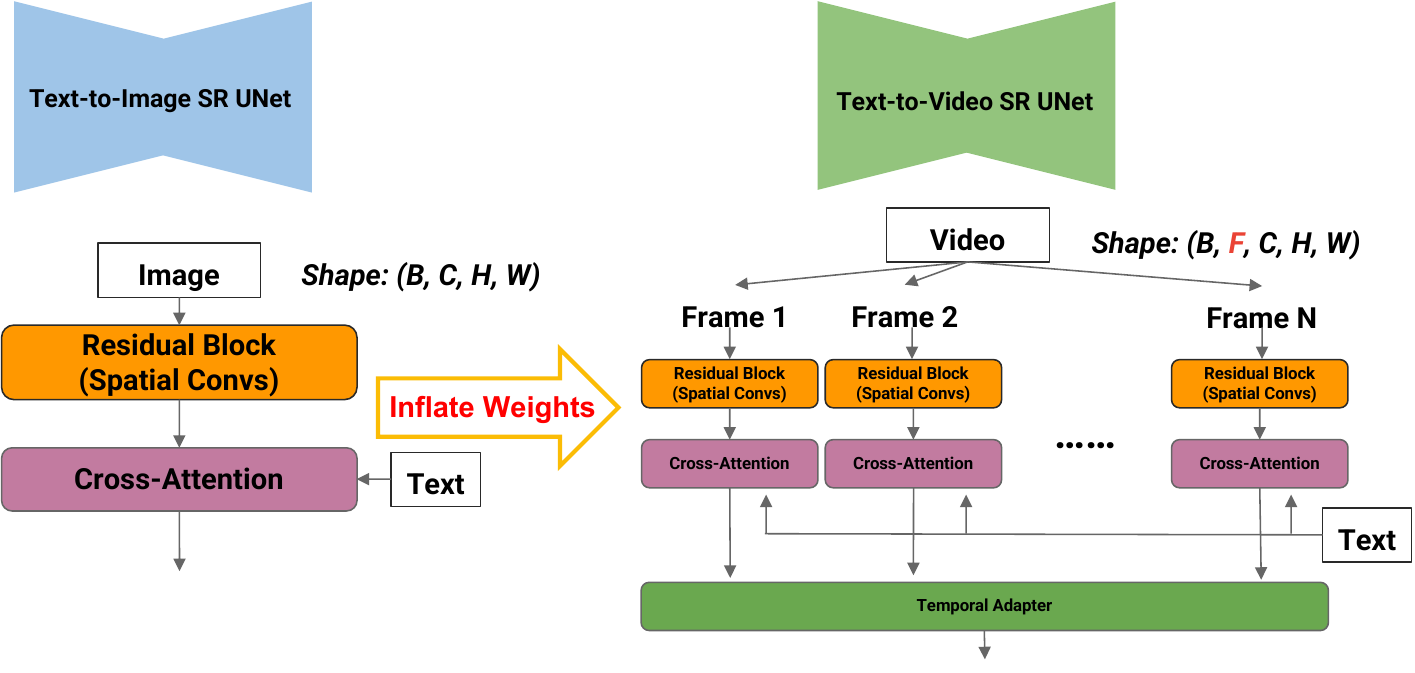}
   \caption{Weights inflation from a text-to-image SR UNet to a text-to-video SR UNet.}
   \label{fig:inflate}
\end{minipage}
\end{figure*}

\begin{figure*}[tbh]
\centering
   \begin{minipage}[t]{\linewidth}
   \centering
   \includegraphics[width=\columnwidth]{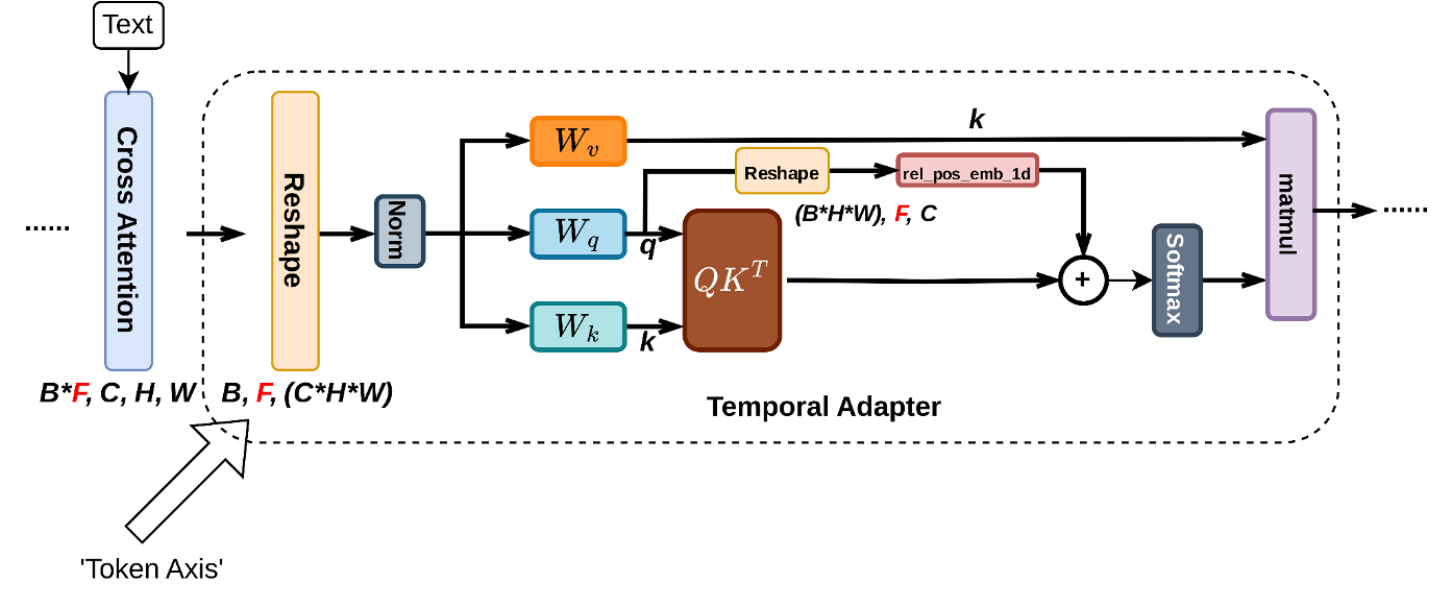}
   \caption{{Temporal adapter with attention that ensures temporal coherence across a video clip.}}
   \label{fig:tp_adapt}
\end{minipage}
\end{figure*}
\begin{figure*}
    \centering
     \includegraphics[width=\linewidth]{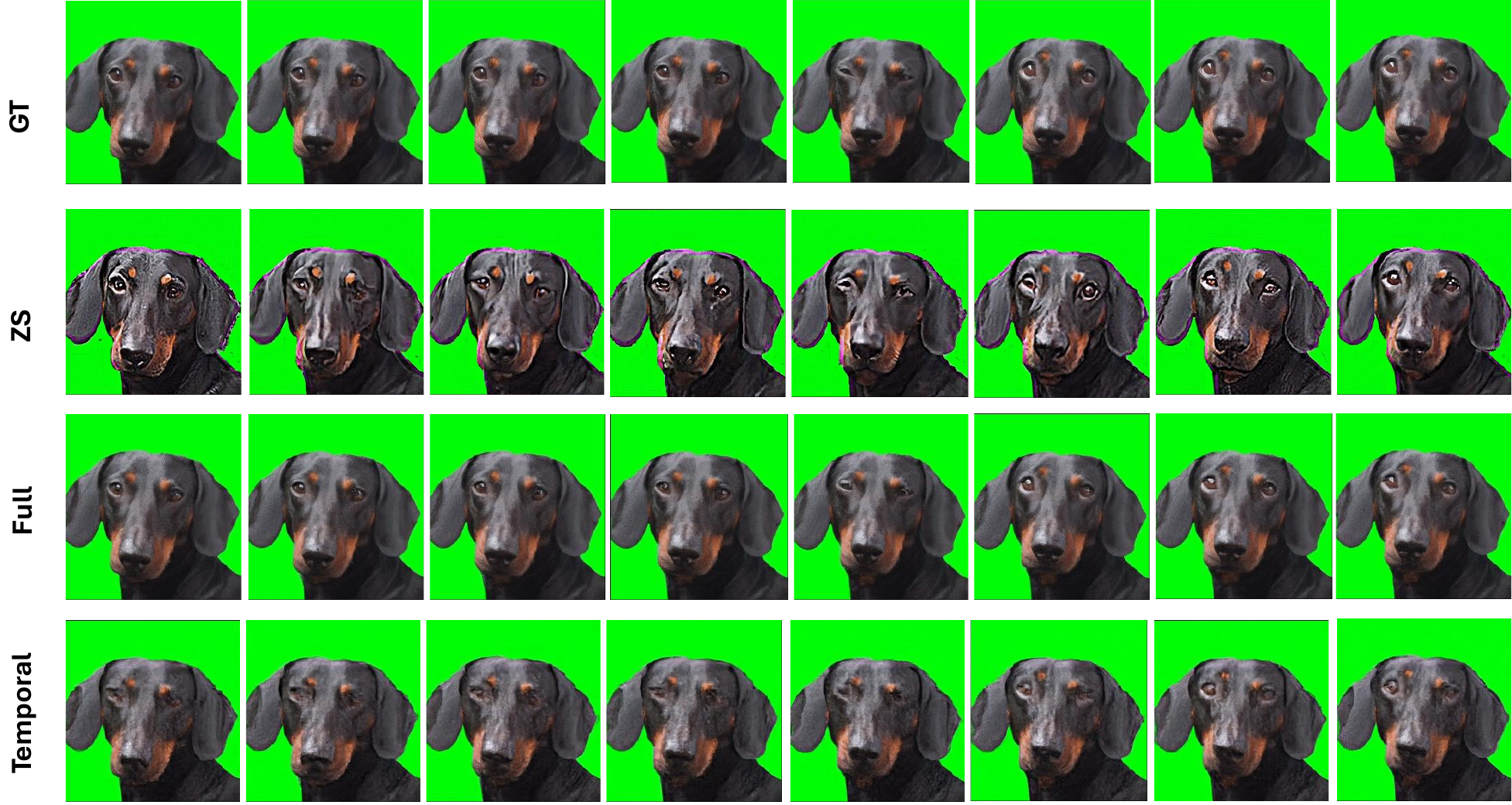 }
    \caption{Visualization of different tuning methods after image model inflation, conditioned on text prompt ``Dog dachshund on chromakey''.}
\label{fig:sr_vis_1}
\end{figure*}

\begin{figure*}[tbh]
    \includegraphics[width=\linewidth]{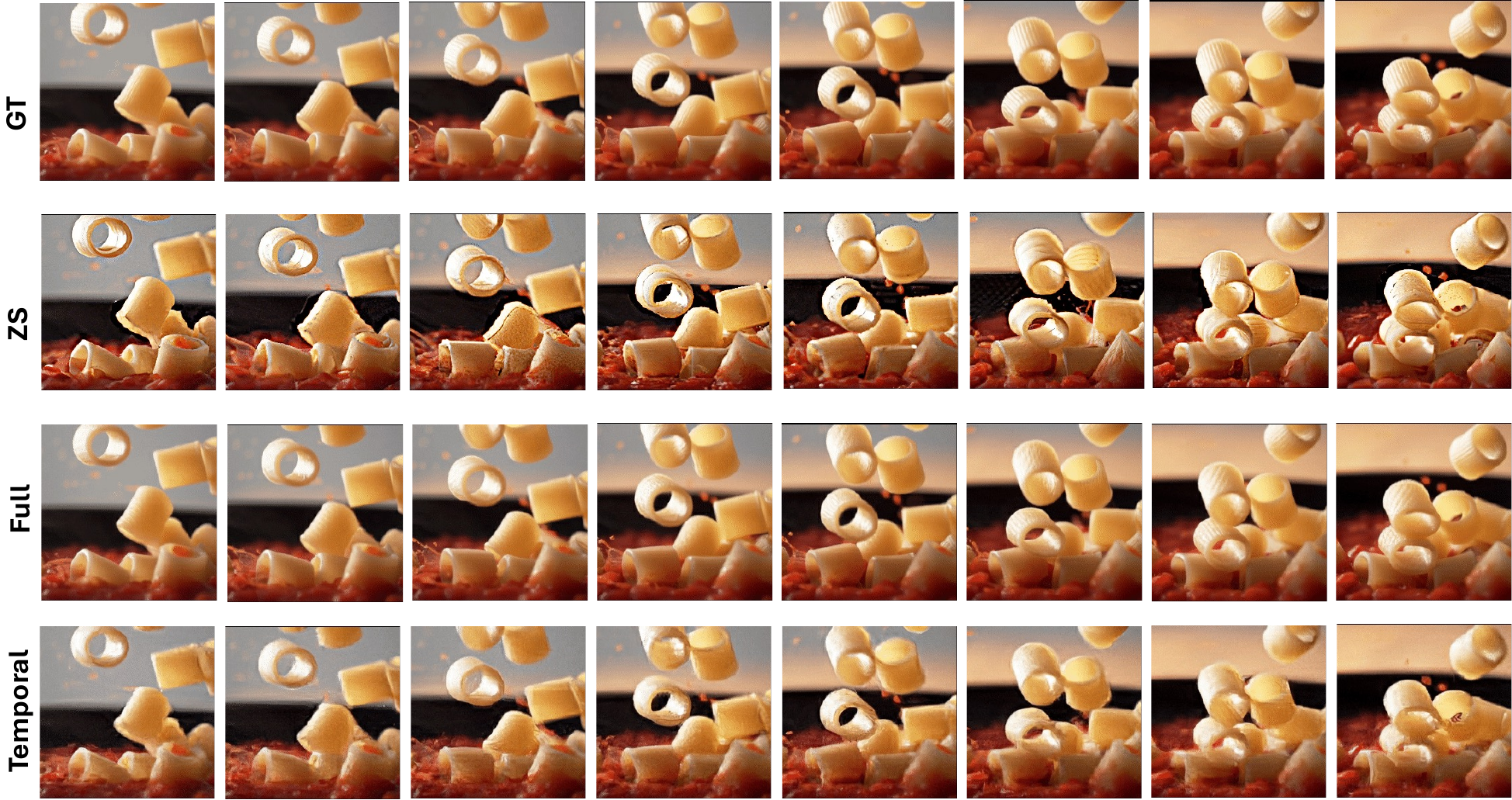}
\caption{Text prompt: Camera follows cooking mezze machine rigate pasta in tomato sauce.}%
\label{fig:sr_vis_more_1}
\end{figure*}

\begin{figure*}[tbh]
    \includegraphics[width=\linewidth]{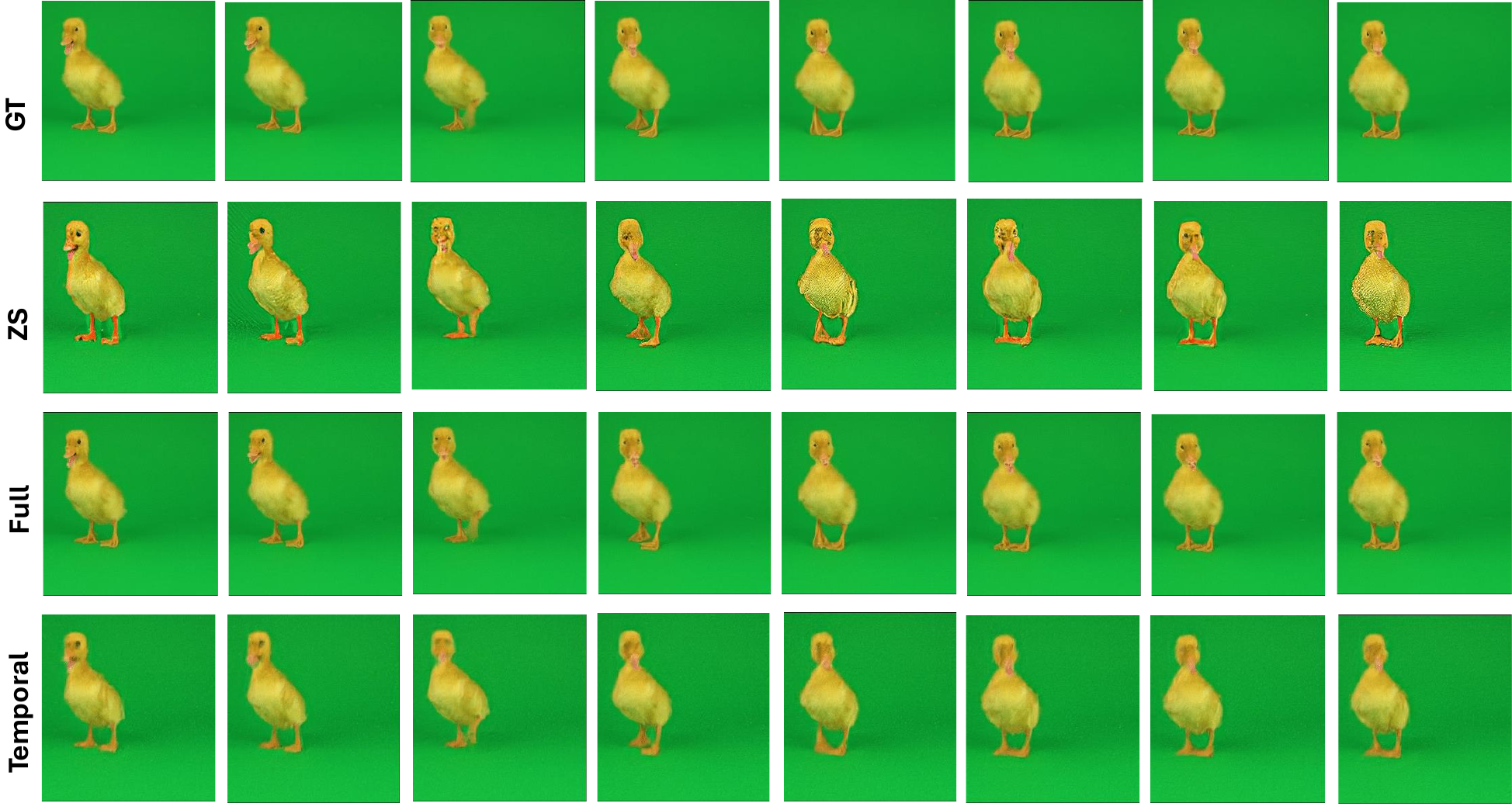 }
\caption{Text prompt: Little beautiful ducklings on green screen.
}%
\label{fig:sr_vis_more_2}
\end{figure*}
\begin{figure*}[tbh]
    \includegraphics[width=\linewidth]{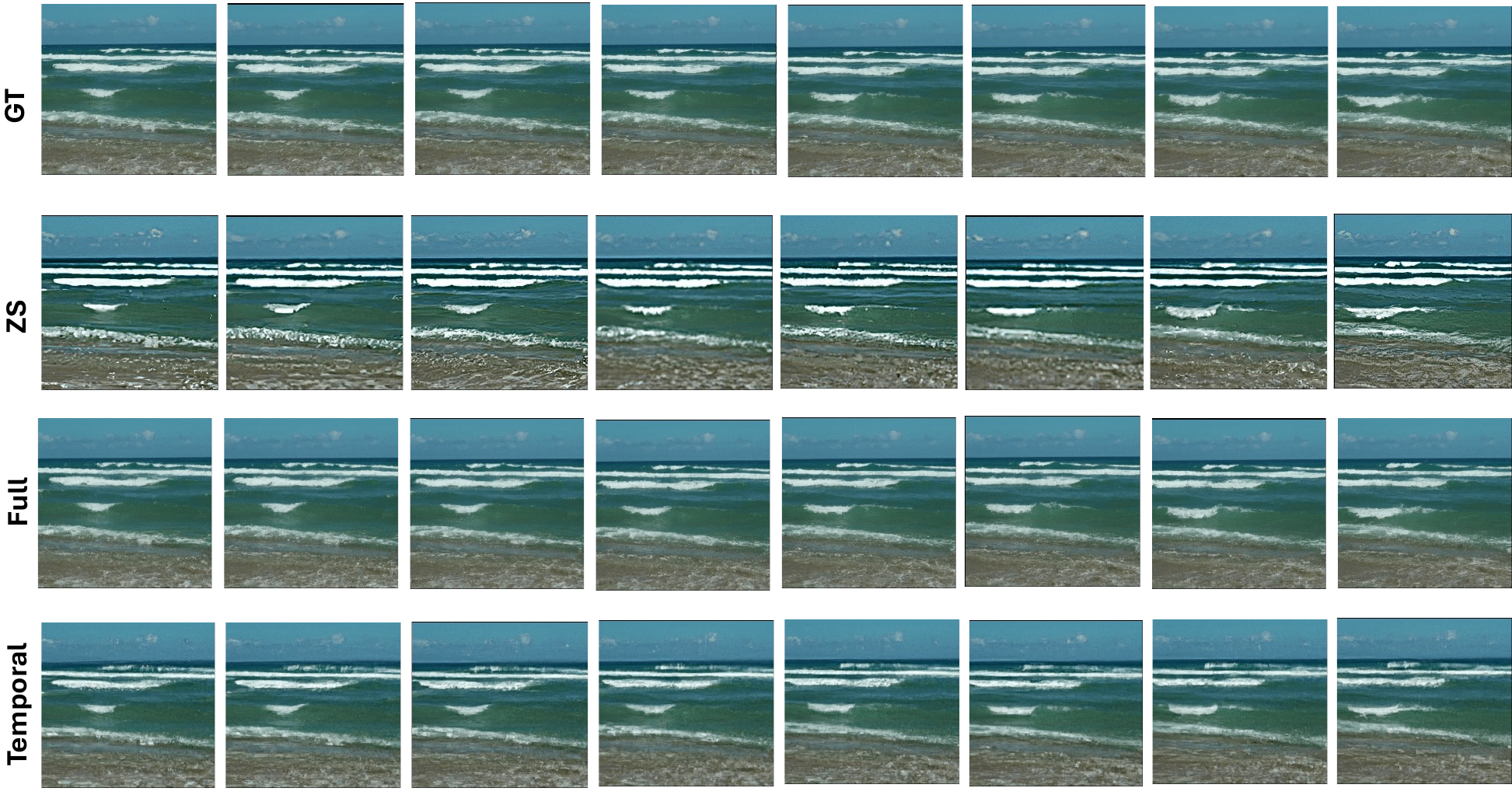 }
\caption{Text prompt: Brazil northeast beaches.}
\label{fig:sr_vis_more_3}
\end{figure*}

\begin{figure*}[tbh]
  \begin{center}
    \begin{minipage}[b]{0.9\linewidth}
\subfigure[Ground Truth]{
    \label{fig:app:exist}
    \centering
    \includegraphics[width=\columnwidth]{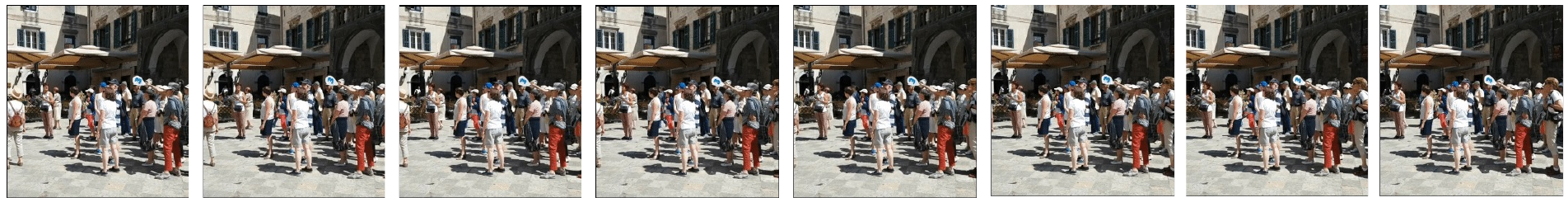 }
}
\subfigure[Train with 40 $\%$ video data]{
    \label{fig:app:ours}
    \centering
    \includegraphics[width=\columnwidth]{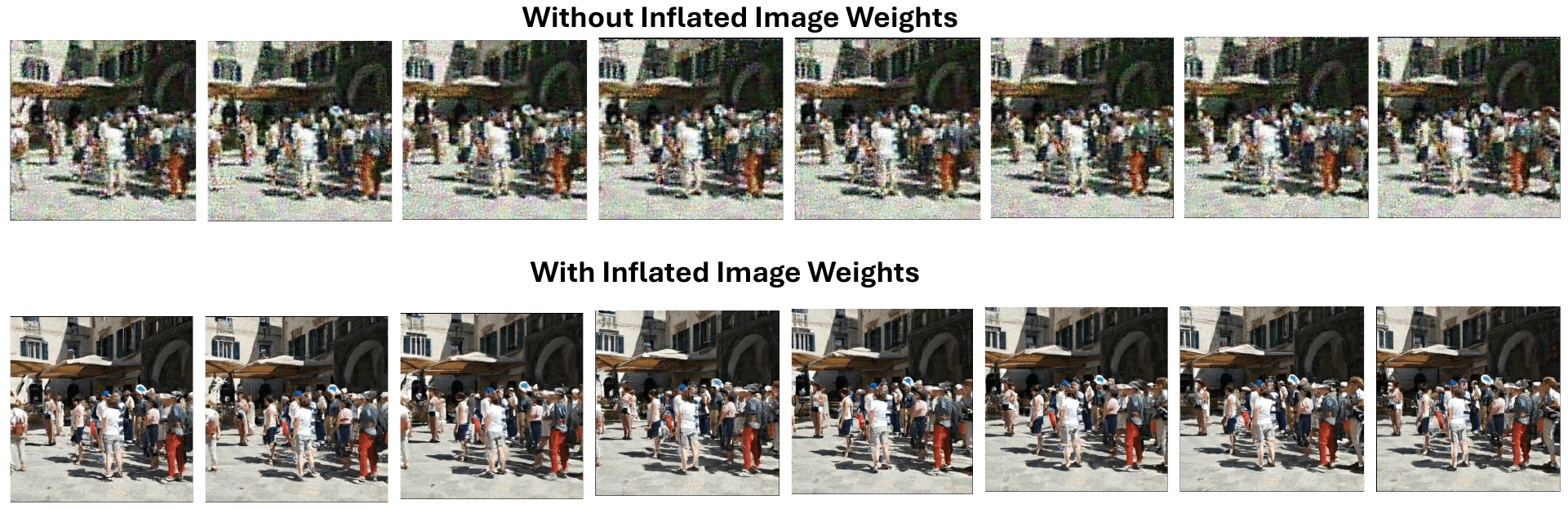}
}
\subfigure[Train with 50 $\%$ video data]{
    \label{fig:app:ours}
    \centering
    \includegraphics[width=\columnwidth]{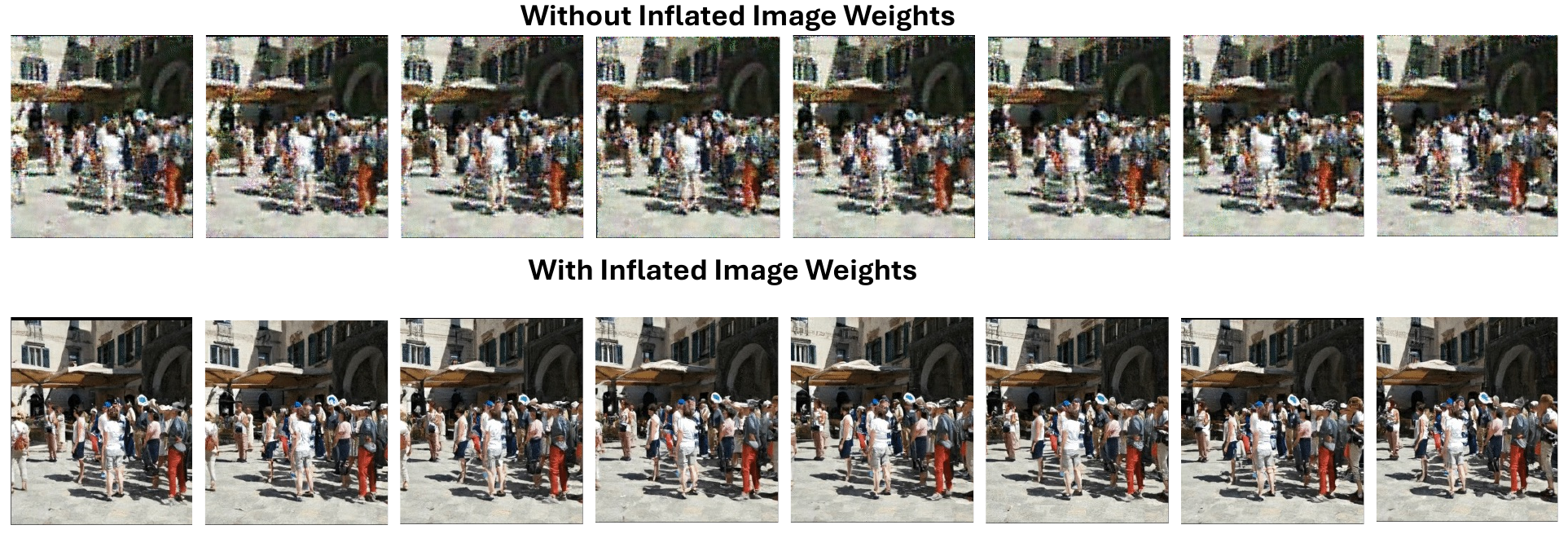}
}
\subfigure[Train with 60 $\%$ video data]{
    \label{fig:app:ours}
    \centering
    \includegraphics[width=\columnwidth]{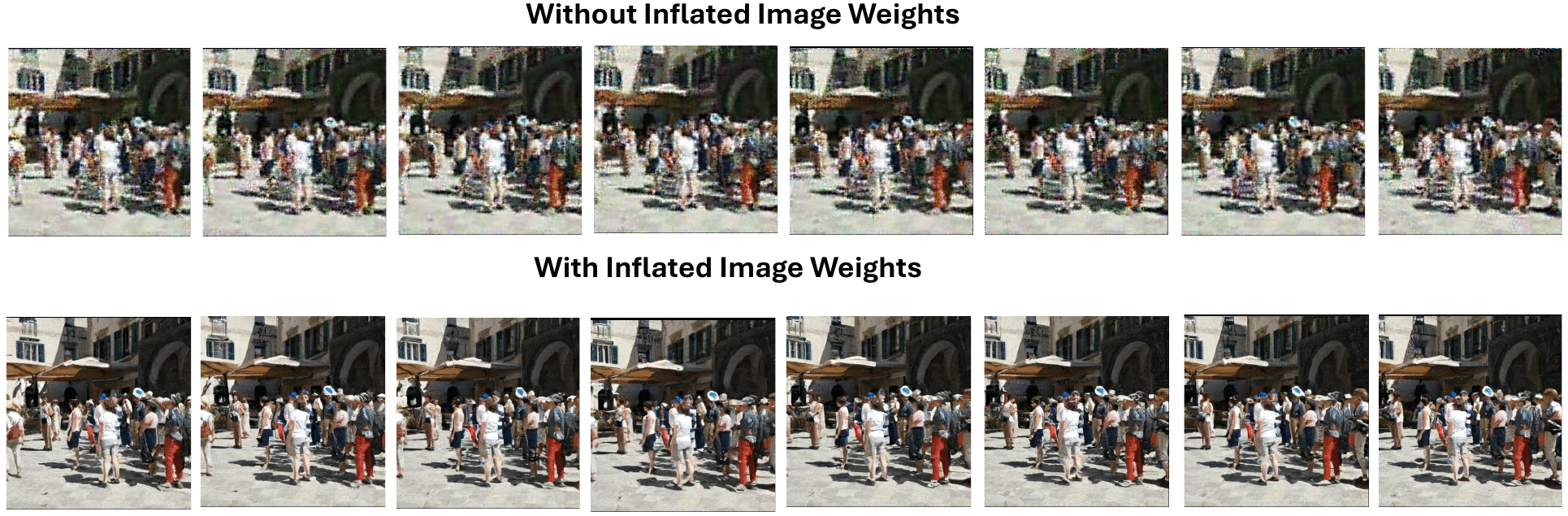}
}
\caption{Visualizations of methods with and without image model inflation. Text prompt: Tourists visiting the old town.
}%
\label{fig:de_vis_2}
\end{minipage}
\end{center}
\end{figure*}
\subsection{Inflation with Image Weights} \label{subsec:inflate}
We build a one-on-one mapping between image and video architectures through `upgrade' Imagen~\cite{DBLP:conf/nips/SahariaCSLWDGLA22} text-to-image super-resolution model to accommodate video tasks.
We first revisit the U-Net architecture in Imagen, composed of residual and cross-attention blocks, as shown in Figure~\ref{fig:inflate} (left).
Given a batch of input static images with shape $B\times C \times H \times W$, the residual blocks capture the spatial information while cross-attention ensures that the generated image aligns with the given text prompt.
In the context of our text-to-video super-resolution, the input batch of video clips is in the shape of $B \times F \times C \times H \times W$, where $F$ is the number of frames. As shown in Figure~\ref{fig:inflate} (right), each individual frame is processed through a parallel scheme, each branch contains a residual block and cross-attention layer for text-visual feature extraction. At the end of the UNet block, we have a temporal adapter for feature aggregation to maintain consistency and smoothness across frames. The processing units of the residual block and cross-attention layer share the same weights during training, in which case we can simply reshape the video data into $(BF) \times C \times H \times W$. Given this nice property of weight sharing scheme, we can directly inflate the pre-trained image model weights into the video UNet without any architectural modification. 

\subsection{Temporal Adaptater with Frame-wise Attention}\label{subsec:adapter}
To capture the coherence among video frames, we apply the temporal adapter after the residual and cross-attention blocks. Figure~\ref{fig:tp_adapt} depicts the design of the attention-based temporal adapter. We first reshape video data $I$ into $I'$ with the shape of $B \times F \times (CHW)$ and then adopt a conventional self attention module:
\begin{eqnarray}\label{eq:attention}
    \text{Self-Attention}(Q,K,V) = \text{Softmax}(\frac{QK^T}{\sqrt{d}}) \cdot V. 
\end{eqnarray}
Such an attention mechanism is effective in determining the overall
structure and the coherence of the video frames. Specifically, a weighted sum over the `token axis' $F$ is calculated to learn the frame-wise correlation.
We employ end-to-end optimization of either the full or partial model weights, aligning with the simple denoising objective in DDPM~\cite{DBLP:conf/nips/HoJA20}. As such, the model weights are optimized by minimizing the MSE loss of noise prediction, conditioned on the low-resolution frames.
\section{Experiments}
We validate our approach on the Shutterstock dataset.
We inflate a version of the Imagen diffusion model pre-trained on our internal data sources for 8x super-resolution, into our video UNet. This UNet consists of four stages each for downsampling and upsampling, denoted as $2\times$, $4\times$, $8\times$, $16\times$.
T5-xxl encoder~\cite{DBLP:journals/jmlr/RaffelSRLNMZLL20} is used to extract text embedding, the output of which is fed into the cross-attention layers within the $16\times$ stage.
We train our video model on the Shutterstock text-to-video dataset with 7 million video clips in a resolution of $256\times 256$-resolution and frame rate of 8 FPS. 
The duration for each clip is 1 second, i.e. $F=8$. The supper-resolution scale is $4\times$, elevating the resolution from $64\times 64$ to $256 \times 256$.
We investigate several baseline optimization approaches, including (1) Zero-shot (ZS): we directly evaluate the video model after inflation without further training. (2) Full-ft (Full): After integrating the temporal adapter, all modules undergo optimization. This strategy aims to showcase the potential 'upper bound' performance in the super-resolution task. (3) Temporal: we only tune the temporal adapter to capture the temporal consistency while maintaining superior generation quality efficiently.
We finetune for 1 epoch, using Adafactor~\cite{DBLP:conf/icml/ShazeerS18} with initial LR of $10^{-5}$ and batch size of 256 on 64 TPUv3.
\begin{table*}
\caption{{Quantitative results for different tuning approaches.}}
\setlength{\tabcolsep}{2pt}
\label{tab:quant:result}
\centering
\begin{tabular}{@{}llccccccccccccccc@{}}
\toprule
\multicolumn{2}{c}{}                        && \multicolumn{2}{c}{Visual Quality} && \multicolumn{1}{c}{Temporal Consistency} && \multicolumn{3}{c}{Efficiency}   \\    [-2pt]  \cmidrule{3-5} \cmidrule{7-8} \cmidrule{9-11}
\multicolumn{2}{c}{\multirow{1}{*}{Method}} &&    PSNR (\% $\uparrow$)      & \multicolumn{1}{c}{SSIM ($\uparrow$)}             &&  \multicolumn{1}{c}{TCC ($\uparrow$)} && Tunable Params (M) $\downarrow$       & \multicolumn{1}{c}{Train Speed (steps/s) $\uparrow$} & \multicolumn{1}{c}{Memory} (G) $\downarrow$ \\[-1pt]
\midrule
Zero-shot    &     &   &  18.1 & 0.42    && 0.70      & &-    & - &-  \\
Full-ft    &     &   &  28.7 & 0.77     && 0.86     & &628.89    &1.05 &15  \\
Temporal    &     &   &  24.3 & 0.62    && 0.82     & &67.24    &2.02 &8  \\
\bottomrule
\end{tabular}
\end{table*}
\subsection{Quantitative Results}
We evaluate different optimization approaches using various metrics.
As shown in Table~\ref{tab:quant:result}, the Full-ft approach achieves the best visual quality in terms of Peak signal to noise ratio (PSNR) and structural index similarity (SSIM) by tuning all $628.89$ million parameters of UNet. The efficient temporal adapter tuning still yields reasonable visual quality while achieving an approximate $2\times$ wall-clock training acceleration and halving memory usage by adjusting only one-tenth of the typical parameter quantity. The zero-shot approach performs the worst.

We also validate that the efficient tuning approach can maintain temporal consistency, i.e. the motions among constructive
frames remain smooth in the super-resolution results. We adopt the quantitative evaluation metric in~\cite{DBLP:conf/iccv/Zhang0C0SY19}: temporal change consistency (TCC), which is defined as:
\begin{equation}
    TCC(H, G) = \frac{\sum_{i=1}^{n-1}SSIM(|h^i-h^{i+1}|,|g^i-g^{i+1}|)}{n-1}
\end{equation}
where $H=\{h^1, h^2, ..., h^n\}$ and $G=\{g^1, g^2, ..., g^n\}$ are high-resolution ground-truth and generated video frames, respectively. Table~\ref{tab:quant:result} shows a clear trade-off between training efficiency and temporal consistency, in which efficient temporal tuning still yields reasonable results. We also observe that zero-shot approach fails to maintain the consistent changes among adjacent frames due to the lack of a temporal module that operate exclusively on the time axis.

\subsection{Qualitative Results}

As shown in Figure~\ref{fig:sr_vis_1}, when compared with the ground truth high-resolution video (GT), both Full and Temporal produce good super-resolution results, marked by high visual quality and temporal smoothness. The ZS approach manages to generate frames with decent visual content without any fine-tuning on video data, but it falls short in maintaining temporal coherence — a limitation due to its pre-training solely on static images.
This demonstrates the effectiveness of our temporal adapter in capturing the coherence across video frames. We provide addtional visualizations in Figure~\ref{fig:sr_vis_more_1}, ~\ref{fig:sr_vis_more_2} and ~\ref{fig:sr_vis_more_3}.

\subsection{Inflation is Data Efficient}
\begin{figure}[thb]
  \begin{center}
    \begin{minipage}[b]{\linewidth}
\subfigure[PSNR]{
    \label{fig:de:psnr}
    \centering
    \includegraphics[width=\columnwidth]{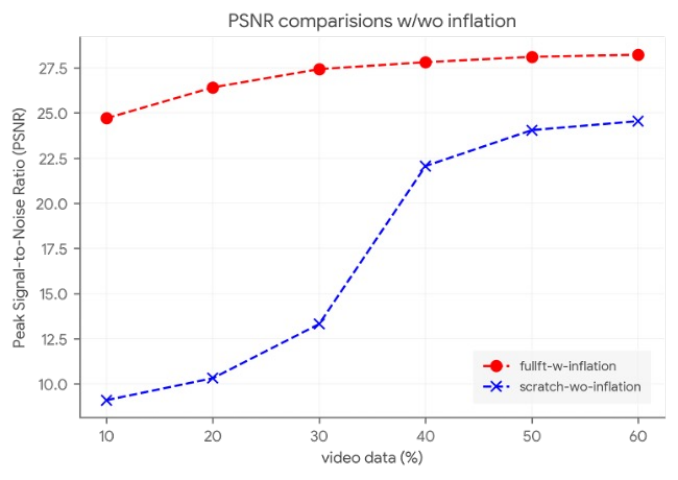}
}
\subfigure[SSIM]{
    \label{fig:de:ssim}
    \centering
    \includegraphics[width=\columnwidth]{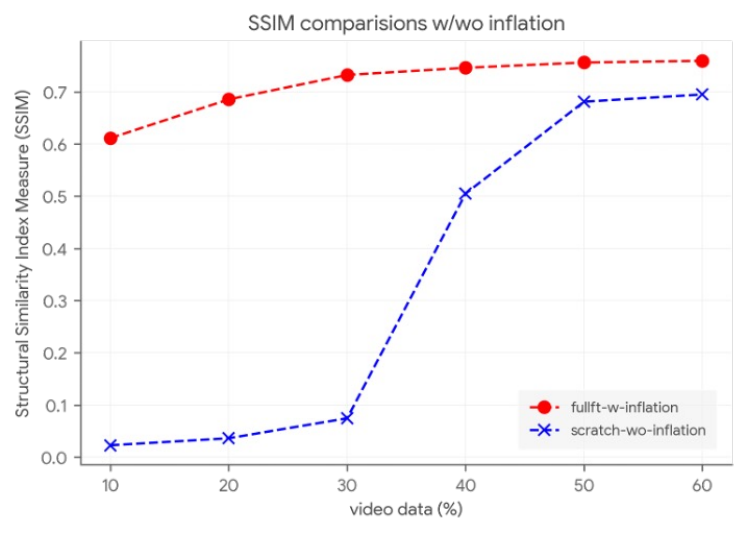}
}
\caption{Training data efficiency evaluated by PSNR and SSIM.
}%
\label{fig:de}
\end{minipage}
\end{center}
\vspace{-1em}
\end{figure}

A straightforward baseline for image weight inflation is to randomly initialize the video UNet and fully fine-tune it using only video data.
As observed in Figure~\ref{fig:de}, the image inflation-based approach can achieve high visual quality even when leveraging only $10\%$ of 7M video data. This trend becomes more evident in Figure~\ref{fig:de_vis_2}, demonstrating the data efficiency of the image weight inflation approach.

\section{Conclusion}
In this paper, we proposed a practical diffusion system for inflating text-to-image model weights to text-to-video spatial super-resolution model. This is the first work to study the weight inflation on the pixel level diffusion model.
We have investigated different tuning methods for efficient temporal adaptation.
We also demonstrated a good trade-off between the super-resolution quality with temporal consistency and tuning efficiency.
As a future investigation, we aim to scale up our target resolution from 256 to 512 (e.g. from $4\times$ to $8\times$SR) and generate videos with longer time frames, which would yield a more obvious trade-off between generation quality and computational resources.

\newpage
\small
\bibliography{ref}
\bibliographystyle{ieee_fullname}

\end{document}